\pdfoutput=1

\documentclass[11pt]{article}

\usepackage{EMNLP2022}

\usepackage{times}
\usepackage{latexsym}
\usepackage{multirow}
\usepackage{enumitem}
\usepackage{ragged2e}

\usepackage{url}
\usepackage{caption}
\usepackage{graphicx}
\usepackage{amsmath}
\usepackage{booktabs}
\usepackage{xcolor}
\usepackage{verbatim}

\usepackage{pifont}
\newcommand{\cmark}{\ding{51}}%

\usepackage[T1]{fontenc}
\usepackage[utf8]{inputenc}
\usepackage{microtype}

\usepackage{hyperref}

\title{CRIPP-VQA: Counterfactual Reasoning about Implicit Physical Properties via Video Question Answering}

\author{
    Maitreya Patel \and  Tejas Gokhale \and  Chitta Baral \and Yezhou Yang\\
    Arizona State University\\
    {\small\texttt{\{maitreya.patel, tgokhale, chitta, yz.yang\}@asu.edu}}
    }

\begin{document}
\maketitle
\begin{abstract}
Videos often capture objects, their visible properties, their motion, and the interactions between different objects.
Objects also have physical properties such as mass, which the imaging pipeline is unable to directly capture.
However, these properties can be estimated by utilizing cues from relative object motion and the dynamics introduced by collisions.
In this paper, we introduce CRIPP-VQA\footnote{\url{https://maitreyapatel.com/CRIPP-VQA/}}, a new video question answering dataset for reasoning about the implicit physical properties of objects in a scene.
CRIPP-VQA contains videos of objects in motion, annotated with questions that involve counterfactual reasoning about the effect of actions, questions about planning in order to reach a goal, and descriptive questions about visible properties of objects.
The CRIPP-VQA test set enables evaluation under several out-of-distribution settings -- videos with objects with masses, coefficients of friction, and initial velocities that are not observed in the training distribution.
Our experiments reveal a surprising and significant performance gap in terms of answering questions about implicit properties (the focus of this paper) and explicit properties of objects (the focus of prior work).
\end{abstract}


\section{Introduction}

\begin{figure}[t!]
    \centering
    \includegraphics[width=\linewidth]{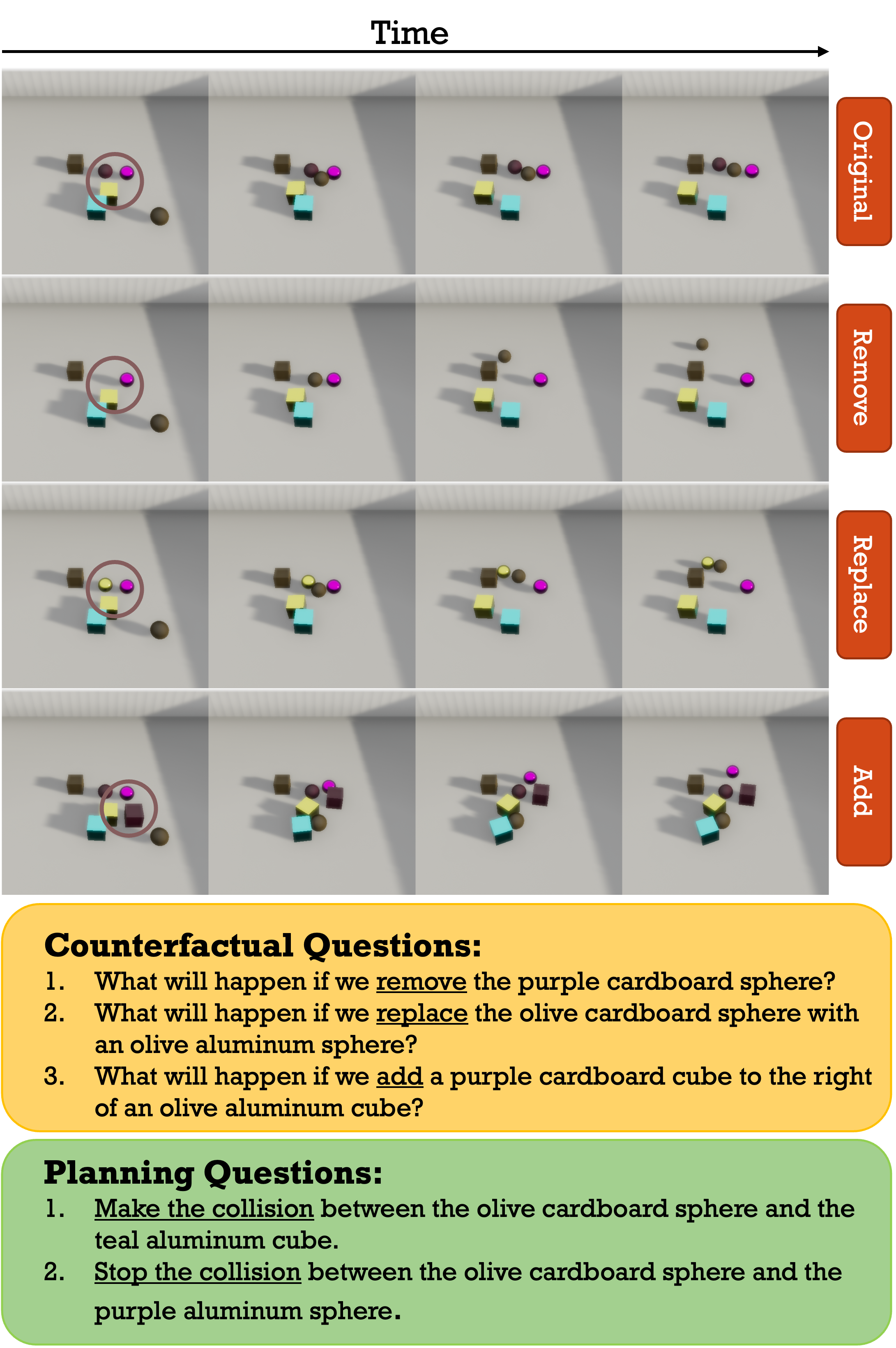}
    \caption{The CRIPP-VQA dataset contains questions about the future effect of actions (such as removing, adding, or replacing objects) as well as planning-based questions. 
    Frames from an example video are shown above with the red highlighted area depicting the objects on which actions (remove, replace, add) are performed.}
    \label{fig:cripp_data}
\end{figure}

Visual grounding seeks to link images or videos with natural language.
Towards this goal, many tasks such as referring expressions~\citep{yu2016modeling}, captioning~\citep{vinyals2015show,xu2016msr}, text-based retrieval~\citep{vo2019composing,rohrbach2015dataset}, and visual question answering~\citep{antol2015vqa,jang2017tgif} have been studied for both images and videos.
Videos often contain objects which can be identified in terms of their visible properties such as their shapes, sizes, colors, textures, and categories.
These visible properties can be estimated by using computer vision algorithms for object recognition, detection, color recognition, shape estimation, etc.
However, objects also have physical properties such as mass and coefficient of friction, which are not captured by cameras.
For instance, given a video of a stone rolling down a hill, cameras can capture the color of the stone and its trajectory – but can they estimate the mass of the stone?
It is therefore difficult to reason about such implicit physical properties, by simply watching videos.

Collisions between objects, however, do offer visual cues about mass and friction.
When objects collide, their resulting velocities and directions of motion depend upon their physical properties, and are governed by fundamental laws of physics such as conservation of momentum and energy.
By observing the change in velocities and directions, it is possible to reason about the relative physical properties of colliding objects.
In many cases, when humans watch objects in motion and under collision, we do not accurately know the masses, friction, or other hidden properties of objects.
Yet, when we interact with these objects, for example in sports such as billiards, carrom, or curling, we can reason about the consequences of actions such as hitting one ball with another, removing an object, replacing an object with a different one, or adding an object to the scene.

In this paper, we consider the task of reasoning about such implicit properties of objects, via the use of language, without having ground truth annotations for the true values of mass and friction of objects. 
We propose a video question answering dataset called CRIPP-VQA, short for \textbf{C}ounterfactual \textbf{R}easoning about \textbf{I}mplicit \textbf{P}hysical \textbf{P}roperties.
Each video contains several objects with at least one object in motion. The object in motion causes collisions and changes the spatial configuration of the scene.
The consequences of these collision are directly impacted by the physical properties of objects.
CRIPP-VQA contains videos annotated with question-answer pairs, where the questions are about the consequences of actions and collisions, as illustrated in Figure~\ref{fig:cripp_data}.
These questions require an understanding of the current configuration as well as counterfactual situations, i.e. the effect of actions such as removing, adding, and replacing objects.
The dataset also contains questions that require the ability to plan in order to achieve certain configurations, for example producing or avoiding particular collisions.
It is important to note that both tasks cannot be performed without an understanding of the relative mass. 
For example, the \textit{``replace''} action can lead to a change in mass inside the reference video, which can drastically change the consequences (i.e., set of collisions). 

We benchmark existing state-of-the-art video question-answering models on the new CRIPP-VQA dataset.
Our key finding is that compared to performance on questions about visible properties (\textit{``descriptive''} questions), the performance on counterfactual and planning questions is significantly low. 
This reveals a large gap in understanding the physical properties of objects from video and language supervision.
Detailed analysis reveals that models can answer questions about the first collision with higher accuracy compared to questions about subsequent future collisions. 

Aloe~\citep{ding2021attention} is a strong baseline for video QA tasks and has improved the state of the art on many previous video QA benchmarks such as CLEVRER~\citep{yi2019clevrer} and CATER~\citep{girdhar2019cater}.
However on CRIPP-VQA, we discovered that the object identification module from Aloe failed to recognize objects in our videos, which we believe is due to the presence of complex textures, reflections, and shadows in our dataset.
To mitigate these failures, we modified Aloe by adapting the Mask-RCNN~\citep{he2017mask} as an object segmentation module.
We also found that using pre-trained BERT-based word embeddings significantly improves the performance over our modified Aloe (Aloe\textsuperscript{*}), serving as the strongest model on CRIPP-VQA.

CRIPP-VQA also allows us to evaluate trained models on out-of-distribution (OOD) test sets, where the objects in videos may have previously unobserved physical properties.
There are four OOD test sets in CRIPP-VQA such that one physical property varies at test time -- objects with a new mass, zero friction coefficient, increased initial velocity, and two moving objects at initialization.
This OOD evaluation reveals a further degradation in performance and a close-to-random accuracy for most state-of-the-art models. 
The results indicate that the scenario with two initially moving objects is the most difficult of all OOD scenarios.

\begin{table*}[!htb]
\centering
\huge
\resizebox{\linewidth}{!}{
\begin{tabular}{@{}lccccccccc@{}}
    \toprule
    \multirow{2}{*}{\textbf{Dataset}} 
        & \multirow{2}{*}{\textbf{\shortstack{Video\\QA}}}
        & \multirow{2}{*}{\textbf{\shortstack{Physical\\Reasoning}}} 
        & \multirow{2}{*}{\textbf{\shortstack{Visually Hidden\\Properties}}} 
        & \multicolumn{3}{c}{\multirow{1}{*}{\textbf{\quad Counterfactual Actions\quad}}}
        & \multirow{2}{*}{\textbf{Planning}}
        & \multirow{2}{*}{\textbf{\shortstack{Physical\\OOD}}} 
        & \multirow{2}{*}{\textbf{\shortstack{Implicit\\Reasoning}}}
        \\
        \cmidrule{5-7}
    & & & & \textbf{Add} & \textbf{Replace} & \textbf{Remove} & & & \\
    \midrule
    MovieQA~\citep{tapaswi2016movieqa} & \cmark & - & - & - & - & - & - & - & - \\ 
    TGIF-QA~\citep{li2016tgif} & \cmark & - & - & - & - & - & - & - & - \\ 
    TVQA/TVQA+~\citep{lei2019tvqa+} & \cmark & - & - & - & - & - & - & - & - \\ 
    AGQA~\citep{Grunde-McLaughlin_2021_CVPR} & \cmark & - & - & - & - & - & - & - & - \\ \midrule
    CoPhy~\citep{baradel2019cophy} & - & \cmark & \cmark & - & - & - & - & - & \cmark \\ 
    CLEVR\_HYP~\citep{sampat-etal-2021-clevr} & - & - & - & \cmark & \cmark & \cmark & -  & - & - \\
    IntPhys~\citep{riochet2018intphys} & \cmark & \cmark & - & - & - & - & \cmark & - & - \\ 
    ESPRIT~\citep{rajani2020esprit} & \cmark & \cmark & - & - & - & - & \cmark & - & - \\ 
    CATER~\citep{girdhar2019cater} & \cmark & - & - & - & - & - & - & - & - \\ 
    CRAFT~\citep{ates2020craft} & \cmark & \cmark & - & - & - & \cmark & -  & - & - \\
    CLEVRER~\citep{yi2019clevrer} & \cmark &  \cmark & - & - & - & \cmark & - & - & - \\ 
    ComPhy~\citep{chen2022comphy} & \cmark & \cmark & \cmark & - & - & - & - & - & - \\ \midrule
    \textbf{CRIPP-VQA \textit{(this work)}} & \cmark & \cmark & \cmark & \cmark & \cmark & \cmark & \cmark & \cmark & \cmark \\ \midrule
\end{tabular}
}
\caption{A comparison of CRIPP-VQA with prior work on video question answering, in terms of different aspects of visual reasoning that are tested.} 
\label{tab:datasets}
\end{table*}

\medskip
\noindent\textbf{Contributions and Findings}:
\smallskip
\begin{itemize}[nosep,noitemsep,leftmargin=*]
    \item We introduce a new benchmark, CRIPP-VQA, for video question answering which requires reasoning about the implicit physical properties of objects in videos.
    \item CRIPP-VQA contains questions about the effect of actions such as removing, replacing, and adding objects, as well as a novel planning task, where the model needs to perform the three hypothetical actions to either stop or make collisions between two given objects. 
    \item Performance evaluation on both \textit{i.i.d.} and out-of-distribution test sets shows the significant challenge that CRIPP-VQA brings to video understanding systems.
\end{itemize}

\section{Related Work}
\label{sec:rel_worl}

\paragraph{Image Question Answering.} 
The VQA dataset~\citep{antol2015vqa} has been extensively used for image-based question answering.
GQA~\citep{hudson2019gqa} and CLEVR~\citep{johnson2016clevr} focus on the compositional and spatial understanding of visual question answering models.
CLEVR-HYP~\citep{sampat-etal-2021-clevr} extends the CLEVR setup with questions about hypothetical actions performed on the image.
OK-VQA~\citep{marino2019ok} deals with answering questions where external world knowledge (such as Wikipedia facts) are required for answering questions, whereas
VLQA~\citep{sampat2020visuo} studies image question answering with additional information provided via an input paragraph.

\paragraph{Video Question Answering.}
Datasets such as MovieQA~\citep{tapaswi2016movieqa}, TGIF~\citep{li2016tgif}, TVQA/TVQA+~\citep{lei2019tvqa+}, and AGQA~\citep{Grunde-McLaughlin_2021_CVPR}, have been introduced for real-world video question answering. 
However, work on video question answering has largely focused on scenes such as movies and television shows.

\paragraph{Physical Reasoning.} 
Visual planning has been explored in \citet{chang2020procedure} and \citet{gokhale2019blocksworld}.
IntPhy~\citep{riochet2018intphys} and ESPRIT~\citep{rajani2020esprit} require reasoning under the influence of gravity.
CATER~\citep{girdhar2019cater} is a video classification dataset, which proposes the challenge of temporal reasoning on actions such as slide, rotate, pick-place, etc.
Recently, the CLEVRER benchmark~\citep{yi2019clevrer} studied the ability to do counterfactual reasoning but only with ``remove'' action. However, all objects in CLEVRER have identical physical properties.
CoPhy~\citep{baradel2019cophy} studied the problem of predicting consequences in the presence of mass as a confounding variable. It does not involve the change in the physical properties during counterfactual reasoning and only studies displacement-based counterfactual object trajectory estimation. 
ComPhy~\citep{chen2022comphy} is a work closest to ours, with the task of learning visually hidden properties in a few-shot setting and performing counterfactual reasoning with a question that explicitly describes the changes in physical properties (\textit{``What if object A was heavier?''}).
In contrast, with three different question categories (descriptive, counterfactual, and planning), our benchmark requires physical properties to be learned from video rather than explicitly expressed in the question.
We position our work in comparison to previous works in Table~\ref{tab:datasets}.

\paragraph{Textual Commonsense Reasoning.} 
PIQA~\citep{bisk2020piqa} is a dataset for physical commonsense reasoning for natural language understanding (NLU) systems . 
CommonsenseQA~\citep{talmor2018commonsenseqa} is a QA dataset that focuses on inferring associated relations of each entity.
Verb Physics~\citep{forbes2017verb} proposes the task of learning relative physical knowledge (size, weight, strength, etc.) for NLU systems.

\paragraph{Visual Commonsense Reasoning.}
VisualCOMET~\citep{park2020visualcomet} is a dataset for inferring commonsense concepts such as future events and their effects from the images and textual descriptions.
Video2Commonsense~\citep{fang2020video2commonsense} is a video captioning task that seeks to include intentions and effects of human actions in the generated caption. 
VCR~\citep{zellers2019recognition} dataset introduces a VQA task that requires commonsense and understanding the scene context in order to answer questions and also to justify the answer.
While,~\citep{sampat2022reasoning} gives the overview of recent advances in multimodal action based reasoning. 

\paragraph{Robustness of Multimodal Models.}
Robustness to distribution shift and language bias has been extensively studied in the VQA domain~\citep{ray2019sunny,gokhale2020vqa,selvaraju2020squinting,kervadec2020roses,li2020closer,agarwal2020towards}.
Shortcuts and spurious correlations have been observed in visual commonsense reasoning~\citep{ye2021case}.
For V+L entailment tasks, \citet{gokhale2022semantically} found that models are not robust to linguistic transformations, while \citet{thrush2022winoground} found that models were unable to distinguish between subject and object of actions.
However most of the work in robust V+L has focused on biases or distribution shift in the language domain.
CRIPP-VQA introduces out-of-distribution evaluation in terms of \textit{physical} properties of objects in a scene.

\section{The CRIPP-VQA Dataset}
CRIPP-VQA, short for Counterfactual Reasoning about Implicit Physical Properties via Video Question Answering, focuses on understanding the consequences of different hypothetical actions (i.e., remove, replace, and add) in the presence of mass and friction as visually hidden properties.

\subsection{Simulation Setup}
\begin{table*}[t]
    \centering
    \small
    \begin{tabular}{@{}lccccc@{}}
        \toprule
        \textbf{Property} &
        \textbf{IID}  &
        \textbf{Mass} & 
        \textbf{Friction} &
        \textbf{Number of objects} &
        \textbf{Velocity}
        \\ \midrule
        
        Shape & (sphere, cube) & - & - & - & - \\
        Color & (purple, teal, olive) & - & - & - & - \\
        Texture & (cardboard, aluminum) & - & - & - & - \\
        Mass & (2, 14) & (2, 8, 14) & - & - & - \\
        Friction & (0.25) & - & (0.0) & - & - \\ 
        \# of moving objects & 1 & - & - & 2 & - \\
        Initial velocity & (14) & - & - & - & (18) \\
        \bottomrule
    \end{tabular}%
    \caption{They key difference between the IID and various OOD evaluation settings in CRIPP-VQA. Here, ``-'' indicates the no change in particular property from the IID setting.
}
\label{tab:property_diff}
\end{table*}
\paragraph{Objects and States.} 

Table \ref{tab:property_diff} summarizes the different properties present in CRIPP-VQA. 
Each object in our dataset has four visible properties: shape (cube or sphere), color (olive, purple, and teal), texture (aluminum and cardboard), and state (stationary, in motion, and under collision). 
Each object also has two invisible properties: mass and coefficient of friction.
Three actions can be performed on each object -- ``remove'', ``replace'', and ``add''.

In this work, we focus on mass and friction as intrinsic physical properties of objects. 
Each unique \textsc{\{shape, color, texture\}} combination is pre-assigned a mass value that is either $2$ or $14$; for instance, all teal aluminum cubes have mass $2$.
Note that these values are not provided as input to the VQA model and need to be inferred in order to perform counterfactual and planning tasks. 
In the training set and \textit{i.i.d.} test set, the coefficient of friction for all objects with the surface is identical and non-zero. For one of the OOD test sets, we make the surfaces and objects frictionless.
Table~\ref{tab:property_diff} shows the object properties for training videos, \textit{i.i.d.} test set and OOD test set.

\paragraph{Video creation.}
We render videos using \textit{TDW}~\citep{gan2020threedworld}. 
In each instance, we initialize the video with either $5$ or $6$ randomly chosen objects, out of which a single object is initialized with a fixed velocity such that it will collide with other objects. 
Here, we keep a constant initial velocity so that the only way to infer mass is through the impact of subsequent collisions.
Each video is $5$ seconds long, with a frame rate of $25$ fps.
We provide annotation and metadata for each video which contains information about object locations, velocities, orientation, and collision at each frame. 
These annotations are further used to generate the different types of question-answer pairs.

\subsection{Question and Answer Generation}
CRIPP dataset focuses on three categories of tasks: 1) Descriptive, 2) Counterfactual, and 3) Planning. 

\paragraph{Descriptive:} These questions involve understanding the visual properties of the scene, including:
\begin{enumerate}[noitemsep,nosep,label={[Type-\arabic*]},leftmargin=*]
    \item Counting the number of objects with a certain combination of visible properties,
    \item Yes/No questions about object types 
    \item Finding the relationship between two objects under collision
    \item Counting the number of collisions
    \item Finding the maximum/minimum occurring object properties.
\end{enumerate}
We do not include questions that require reasoning over mass, to avoid the introduction of spurious correlation which may influence counterfactual and planning-based questions.

\paragraph{Counterfactual.} 
These questions focus on action-based reasoning. 
We generate a hypothetical situation based on one of these actions, and the task is to predict which collisions may or may not happen if we perform the action on an object. 
\textit{``Remove''} action focuses on a counterfactual scenario where a certain object is removed from the original video. 
\textit{``Replace''} action focuses on a counterfactual scenario where one object is replaced with a different object. A key point to note is that the replace action not only changes the object but can also cause a change in an hidden property. 
\textit{``Add''} action-based questions focus on evaluating the system's understanding of spatial relationship along with the hidden properties, where we create a new hypothetical condition by placing a new object to the \textsc{left/right/front/behind} at a fixed distance from the reference object. 

\paragraph{Planning.} 
CRIPP also contains planning-based questions, where the task is predict an action to perform on objects in the given video to either create or avoid collisions. 
Here, the system needs to predict which action has to be performed and on which object, to achieve the goal. 

\begin{figure}[t!]
    \centering
    \includegraphics[width=\linewidth]{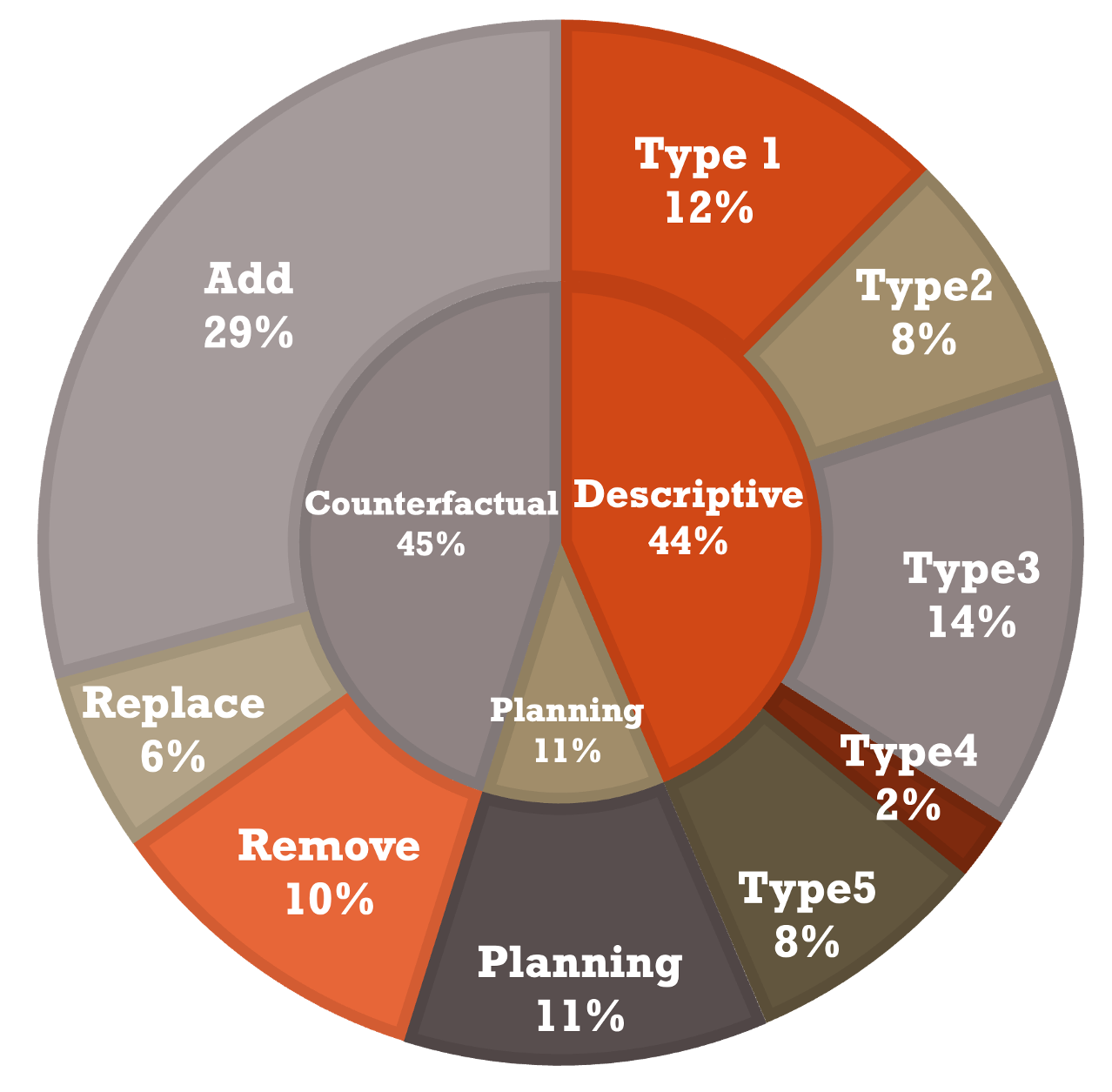}
    \caption{
    A pie-chart showing the distribution of various question types in the CRIPP-VQA dataset. 
    Inner pie chart shows the three broad categories of questions (counterfactual, descriptive, planning),  while the outer pie-chat shows a fine-grained categorization.
    }
    \label{fig:cripp_data_stat}
\end{figure}

\subsection{Dataset Statistics}

\begin{figure*}[t]
    \centering
    \includegraphics[width=0.9\linewidth]{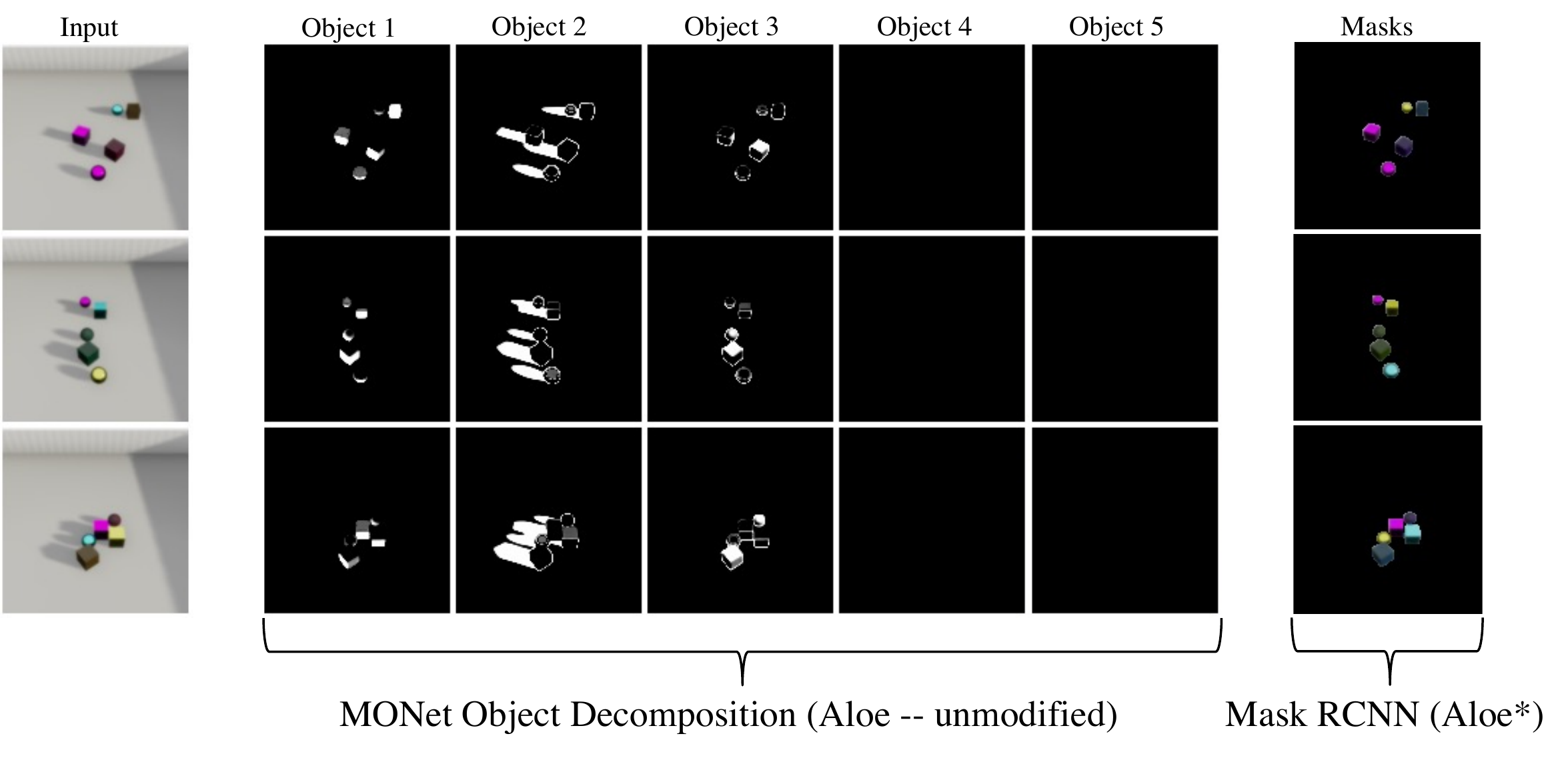}
    \caption{Illustration of the failure of MONet (the object decomposition module in Aloe {\cite{ding2021attention})} on CRIPP-VQA videos. The intended functionality of MONet is to decompose individual objects into separate masks. However as shown above, the predicted masks contain areas corresponding to more than one objects.
    We modified Aloe by replacing MONet with Mask-RCNN, and this approach (Aloe\textsuperscript{*}) leads to more reliable object detection which can be used by the downstream question-answering module.}
    \label{fig:monet_failure_mask}
\end{figure*}

\label{sec:stat}
CRIPP contains $4000$, $500$, and $500$ videos for training, validation, and testing, respectively. Additionally, it has about $2000$ videos focused on evaluation for physical out-of-distribution scenarios. 
CRIPP training dataset has about $41761$ descriptive questions, $41761$ counterfactual questions ($9603$, $5142$, and $27016$ questions for remove, replace, and add actions, respectively), and $10440$ planning-based questions. 
Figure \ref{fig:cripp_data_stat} shows the percentages of each subcategory within the dataset.

\section{Experiments}

\subsection{Problem Statement}
\label{sec:problem}
 Given an input video ($v$), and a question ($q$) the task is to predict the answer ($a$). 
 Each video $v$ contains $m$ number of objects randomly selected from the set $O=\{o_1, o_2, ..., o_n\}$.
 Here, object $o_i$ has several associated properties (i.e., $o_i = (m_i, c_i, s_i, t_i, l_i, v_i)$), where color ($c_i$), shape ($s_i$), texture ($t_i$), location ($l_i$), and velocity ($v_i$) are visually observable properties alongside with mass ($m_i$) as the hidden property.
 More formally, we need to learn the probability density function $F$ such that we maximize the $F(a | v,q)$.

\paragraph{Evaluation Metrics.} 
To evaluate the models, we use two accuracy metrics -- per-option (PO) and per-question (PQ) accuracy. 
Each counterfactual question has multiple options describing the collisions.
Per-option accuracy refers to the option-wise performance and per-question accuracy considers whether all options are correctly predicted or not.
Each planning task involves performing an action over objects within a video. 
Because of this, there may be more than one way to accomplish the objective.
We use \textit{TDW} to re-simulate the models' predictions on the original video to check whether the given planning goal is achieved or not, leading to iterative performance evaluation.

\subsection{Benchmark Models}
We consider three different state-of-the-art models for the video question answering task: 
MAC~\citep{hudson2018compositional}, HCRN~\citep{le2020hierarchical}, and  Aloe~\citep{ding2021attention}.
\textbf{MAC} (Memory, Attention, and Composition) is designed for compositional VQA. 
We modify it by performing channel-wise feature concatenation of each frame, where the channel will contain temporal information instead of spatial information allowing MAC to adapt to the video inputs. 
\textbf{HCRN} (Hierarchical Conditional Relation Network) uses a hierarchical strategy to learn the relation between the visual and textual data. 
\textbf{Aloe} (Attention over learned embeddings) is one of the best-performing models on the CLEVRER~\citep{yi2019clevrer} benchmark. 
It is a transformer-based model, designed for object trajectory-based complex reasoning over synthetic datasets. 
Aloe uses MONet~\citep{burgess2019monet} for obtaining object features by performing an unsupervised decomposition of each frame into objects. 
Aloe takes these frame-wise object features to predict the answers to the input question, using the \textit{[CLS]} token and employs a self-supervised training strategy.

\paragraph{Drawbacks of Aloe.}
We found that the MONet module used in Aloe is very unstable and fails to produce reliable frame-wise features on videos from CRIPP. 
MONet is not able to recognize simple object properties such as color and is not able to decompose the image into masks corresponding to individual objects.
This drawback hurts the performance of Aloe on the CRIPP-VQA dataset, even though Aloe is one of the best-performing models on previous video QA benchmarks.
An example is shown in Figure~\ref{fig:monet_failure_mask}, and more details can be found in Appendix \ref{sec:monet-appendix}.
We believe that this failure could be a result of shadows and textures in our dataset that are not found in previous datasets.

\begin{table*}[t]
\centering
\small
\resizebox{\textwidth}{!}{%
\begin{tabular}{@{}l c cc cc cc c c@{}}
\toprule
\multirow{2}{*}{\textbf{Model}} &  \multirow{2}{*}{\textbf{Descriptive}} 
&  \multicolumn{2}{c}{\textbf{Remove}} & \multicolumn{2}{c}{\textbf{Replace}} & \multicolumn{2}{c}{\textbf{Add}} & \textbf{Counterfactual} & \multirow{2}{*}{\textbf{Planning}} \\
 & & PQ & PO & PQ & PO & PQ & PO & Avg. PO & \\
 \midrule
Frequency & 8.21 & 0.00 & 50.18 & 0.00 & 50.00 & 0.00 & 50.00 & 50.06 & 3.49 \\
Random & 8.51 & 7.21 & 49.58 & 3.34 & 49.40 & 9.39 & 50.04 & 49.67 & 7.39 \\
Blind-BERT & 53.82 & 20.18 & 54.67 & 17.57 & 50.45 & 15.86 & 51.55 & 52.22 & 8.11 \\ \midrule 
MAC~\cite{hudson2018compositional} & 48.72 & 16.41 & 50.68 & 17.31 & 50.21 & 16.29 & 49.83 & 50.24 & 6.26 \\
HCRN~\cite{le2020hierarchical} & 64.98 & 27.20 & 59.04 & 19.87 & 55.97 & 20.49 & 56.06 & 57.02 & 21.38 \\
Aloe\textsuperscript{*} & 68.94 & 31.10 & 62.90 & 9.91 & 52.10 & 18.13 & 56.55 & 57.18 & 31.76 \\
Aloe\textsuperscript{*}+BERT & 71.04 & 33.64 & 65.46 & 22.07 & 56.76 & 39.71 & 67.43 & 63.21 & 32.61 \\ \bottomrule 
\end{tabular}%
}
\caption{
Results on the \textit{i.i.d.} test set showing performance of models evaluated in terms of per-question (PQ) accuracy and per-option (PO) accuracy. For descriptive and planning questions, only one of the answer options are true, therefore per-question and per-option accuracies are identical. 
Aloe\textsuperscript{*} refers to our modified Aloe, where we replace the MONet module with a Mask-RCNN object detector.
}
\label{tab:benchmark}
\end{table*}
\paragraph{Modifying Aloe.}
Due to the failures of the MONet object decomposition module, the Aloe baseline fails measurably on CRIPP-VQA, exhibiting close-to-random performance. 
Therefore, we propose additional modifications to Aloe to make it more widely applicable beyond prior datasets that are built using the CLEVR~\citep{johnson2016clevr} rendering pipeline. 
First, we replace MONet with Mask-RCNN~\citep{he2017mask} to perform instance segmentation and then train an auto-encoder to compress the mask-based object-specific features to make it compatible with Aloe. 
Second, instead of learning the word embedding from scratch, we further propose to use pre-trained BERT-based word embeddings as an input to the Aloe, which leads to a faster and more stable convergence. 
Further architecture modifications and hyper-parameter settings are in Appendix \ref{sec:training}.

We also consider a \textit{``random''} baseline which randomly selects one answer from a possible set of answers, and a \textit{``frequent''} baseline which always predicts the most frequent label.
To analyze textual biases, we use a pretrained text-only QA model (BERT~\citep{devlin2018bert}) that takes only questions as input to predict the answer and ignores the visual input. We denote it as \textit{``Blind-BERT''}.

\subsection{Results}
Table \ref{tab:benchmark} summarizes the performance comparisons of our baselines on the CRIPP-VQA \textit{i.i.d.} test set.
On \textbf{Descriptive} questions, the \textit{``random''} and \textit{``frequent''} baselines achieve around only $8\%$ accuracy, while Blind-BERT gets 53.82\% which suggests the existence of language bias associated with correlations between question types and most likely answers for each.
Surprisingly, MAC achieves only 48.72\% which is lower than Blind-BERT. 
This implies that the video feature representations learned by MAC hurt performance compared to text-only features.
An unmodified version of the Aloe also achieves only ~56\% accuracy.
HCRN and both Aloe variants (Aloe\textsuperscript{*} and Aloe\textsuperscript{*}+BERT) improve performance indicating that visual features are crucial for descriptive questions.
Aloe\textsuperscript{*}+BERT is the best-performing model which implies that our modification with BERT embeddings helps.

\textbf{Counterfactual} questions involve a total of three types of actions. Table (\ref{tab:benchmark}) shows the action-wise performances.
MAC's performance is once again comparable to Blind- BERT's.
HCRN performs slightly better than Blind-BERT. 
This shows that even though visual features in HCRN are better than MAC but it is not sufficient enough to perform such complex reasoning. 
While, unmodified Aloe achieves an average accuracy of $ \sim $ 52\% on counterfactual questions, which is close-to-random performance.
Aloe\textsuperscript{*}+BERT achieves much better results only in terms of remove and add actions.
However, Aloe\textsuperscript{*}+BERT is close to random for questions with the \textit{``replace''} action
as it directly involves the change in physical properties (i.e., mass and shape) of an existing object within the given scenario. 
This implies that Aloe\textsuperscript{*}+BERT is able to perform spatial reasoning to some extent, but is not good at reasoning about changes in physical properties.
As Aloe\textsuperscript{*}+BERT outperforms Aloe\textsuperscript{*} across all actions, it can be implied that BERT-based embeddings enable the model to learn the relationship between objects and actions.

\textbf{Planning-based questions} can have multiple possible answers.
We observe a similar trend in results as Aloe\textsuperscript{*}+BERT performs better than the other baselines.
Further analysis on Aloe\textsuperscript{*}+BERT predictions shows that model predicts ``remove'', ``replace'', and ``add'' actions for planning tasks with 70.52\%, 10.6\%, and 18.87\%, respectively.
This tells us that the model finds it easy to reason when ``remove'' hypothetical action is present.

\paragraph{Human evaluations.} 
We conduct a small-scale human study to gauge an estimate for human-level performance on the CRIPP-VQA dataset.
We had $n=6$ participants in the study.
The participants were habituated to the task by showing them 5 videos and corresponding QA pairs.
Then we asked them to answer 30 questions on different sets of randomly selected videos.
The results of the human study indicate that human participants achieved $90.00\%$, $78.89\%$, and $58.87\%$ on descriptive, counterfactual, and planning tasks, respectively.

\begin{figure}
    \centering
    \includegraphics[width=\columnwidth]{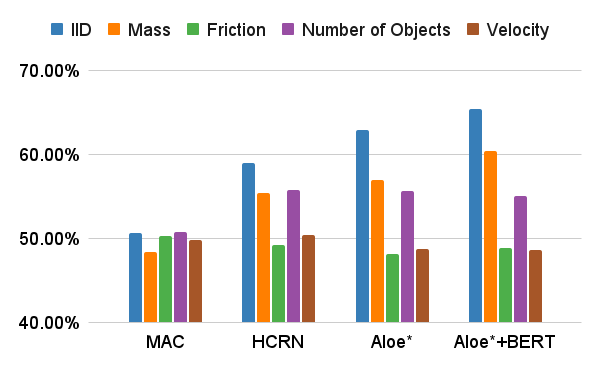}
    \caption{Comparison of performance of models (per-option accuracy) for ``remove'' questions when tested using the IID test set and each OOD test set.  }
    \label{fig:remove_ood}
\end{figure}

\begin{figure}[t]
    \centering
    \includegraphics[width=\columnwidth]{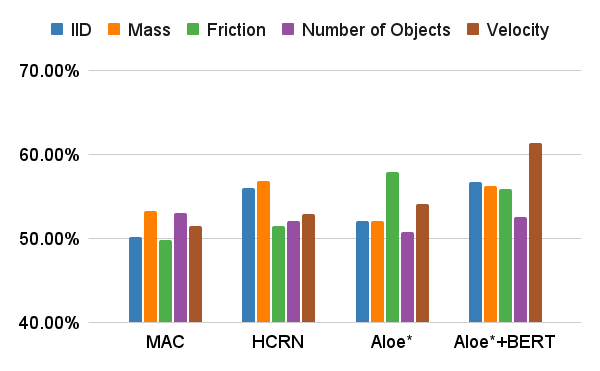}
    \caption{Comparison of performance of models (per-option accuracy) for ``replace'' questions when tested using the IID test set and each OOD test set.}
    \label{fig:replace_ood}
\end{figure}

\begin{figure}[t]
    \centering
    \includegraphics[width=\columnwidth]{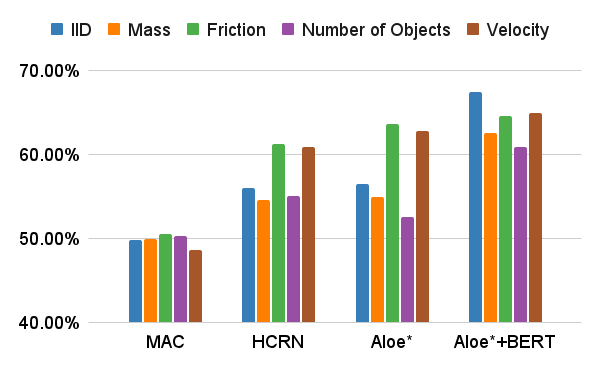}
    \caption{Comparison of performance of models (per-option accuracy) for ``add'' questions when tested using the IID test set and each OOD test set. }
    \label{fig:add_ood}
\end{figure}

\begin{figure}[t]
    \centering
    \includegraphics[width=\columnwidth]{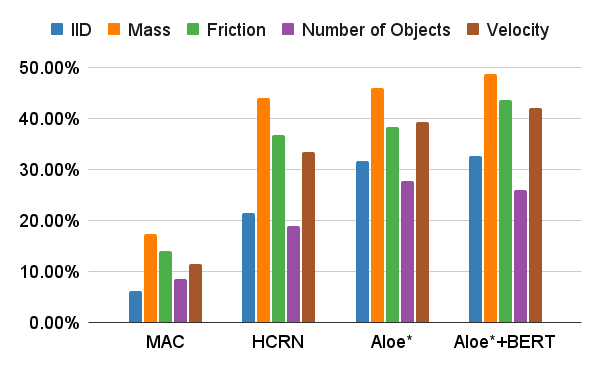}
    \caption{Comparison of performance of models on ``planning'' questions when tested using the IID test set and each OOD test set. }
    \label{fig:planning_ood}
\end{figure}

\subsection{Physical out-of-distribution experiments}
Most of the previous studies focus on feature-based OOD cases (like the rotation of the entities within the image). 
We propose a new dimension of OOD evaluation involving physical properties, by considering four types of OOD scenarios: 
\begin{enumerate}[nosep,noitemsep,leftmargin=*]
    \item \textit{Mass}: the mass of a few objects is changed to 8, 
    \item \textit{Friction}: the surface friction is changed to zero, 
    \item \textit{Number of Objects}: two objects are moving instead of one when the scene is initialized,
    \item \textit{Velocity}: initial object velocity is increased to 18 from 14.
\end{enumerate}

Figures~\ref{fig:remove_ood},\ref{fig:replace_ood},\ref{fig:add_ood},\ref{fig:planning_ood} show the comparison of VideoQA models on \textit{i.i.d.} and different OOD scenarios for \textit{remove}, \textit{replace}, and \textit{add} actions, and planning questions, respectively.
It can be seen that the performance of models becomes close to random ($~50\%$).
This suggests that models are very sensitive to such physical variations at test time, especially for the \textit{``remove''} action (as shown in Figure \ref{fig:remove_ood}). 
From Figure \ref{fig:add_ood}, we observe that the performance drop is negligible across the OOD sets for the add action, especially for Aloe\textsuperscript{*}+BERT. 
Moreover, Figure \ref{fig:planning_ood} shows that the performance increases on several OOD scenarios for planning task. 
At the same time, the performance of bias-check baselines also improves.
This suggests that the expected behavior of the model changes based on the given physical properties.
In the case of the remove action, Friction and Velocity OOD settings are the hardest for all models.
For the \textit{replace} action, the OOD setting with multiple objects is the hardest for Aloe\textsuperscript{*}+BERT.
Number of initial moving Objects based OOD setting is also difficult for models to understand, especially for the add action based questions.

\section{Analysis}
\label{sec:analysis}

In this section, we raise several important questions and derive the insights accordingly.

\paragraph{Performance for detecting present vs. absent collision detection.} 
Consider the example with three objects (A,B,C), where only object A collides with B. 
In this case, we categorize the collision between A \& B as the actual collision ( i.e., prediction label \textit{true}) and the collision between B \& C and A \& C as an absent collision (i.e., prediction label \textit{false}).
Following this rule, we independently check the performance of detecting all occurring collisions and the collisions that never happened. 
Table \ref{tab:true_false} shows the action-based performance of Aloe\textsuperscript{*}+BERT on these two categories. 
It can be inferred that detecting the actual set of collisions is easy except for the \textit{``add''} action, where model mainly predicts that none of the collisions are present in counterfactual scenario. 
However, in the case of the replace action, the model is failing in both categories. 

\paragraph{Performance for First Collision vs Subsequent Collisions.} 
In the CRIPP-VQA dataset, a collision between a pair of objects may lead to subsequent collisions between other objects.
We analyze the performance of the best model (Aloe\textsuperscript{*}+BERT) on counterfactual questions, by comparing the accuracy on questions about the first collision, with the accuracy on questions about subsequent collisions.
To correctly predict subsequent collisions, models need to understand the mass of the objects involved in the first collision to learn the consequences (i.e., sequence of future events). 
From Table (\ref{tab:category_diff}), we observe that for all three actions, there is a drop in performance on subsequent collisions; the drop is highest ($28.48\%$) for \textit{``remove''}.

\paragraph{Importance of mass as intrinsic property.}
There are many hidden factors (i.e., mass, friction, object shape, velocity) that play a role in determining object trajectories and collisions.
To understand these dynamics, we analyze the number of collisions in different counterfactual scenarios and collisions between two different types of objects (in terms of mass).
Table \ref{tab:first_col} shows that if the first collision is between two light or two heavy objects then it leads to almost similar number of collisions.
If the first collision is between light and heavy objects then the number of collisions either decreases or increases.
On average there are $3.00$, $2.06$, $3.31$, and $4.15$ collisions in vanilla, ``remove'', ``replace'', and ``add'' counterfactual settings, respectively.

To summarize, these analyses show that each of our counterfactual scenarios presents unique challenges.
This also strengthens our argument that models fail to learn various reasoning capabilities including but not limited to intrinsic physical properties and consequences of the actions.

\begin{table}[!t]
\centering
\small
\begin{tabular}{@{}lcc@{}}
\toprule
\textbf{Action} &
\textbf{Present collisions}  &
\textbf{Absent collisions} 
\\ \midrule
Remove & 78.27 & 52.81 \\
Replace & 65.74 & 60.23 \\
Add & 46.41 & 79.47 \\ 
\bottomrule
\end{tabular}%
\caption{Per-option accuracy of Aloe\textsuperscript{*}+BERT for detecting present vs absent collisions.}
\label{tab:true_false}
\end{table}
 \begin{table}[!t]
\centering
\small
\begin{tabular}{@{}lccc@{}}
\toprule
\multirow{2}{*}{\textbf{Action}} 
    & \multirow{2}{*}{\textbf{\shortstack{First\\Collision}}} 
    & \multirow{2}{*}{\textbf{\shortstack{Subesequent\\Collisions}}} 
    & \multirow{2}{*}{\textbf{Difference}}\\
\\ \midrule
Remove & 90.52 & 62.45 & 28.07 \\
Replace & 75.38 & 66.03 & 9.35 \\
Add & 55.45 & 41.01 & 14.44 \\ \bottomrule

\end{tabular}%
\caption{Per-option accuracy of Aloe\textsuperscript{*}+BERT for detecting first collision vs. subsequent collisions from the set of occurring collisions in counterfactual scenario.
}
\label{tab:category_diff}
\end{table}
\begin{table}[!t]
\centering
\small
\resizebox{\columnwidth}{!}{%
\begin{tabular}{@{}lcccc@{}}
\toprule
\textbf{First collision type} &
\textbf{L $\rightarrow$ L} &
\textbf{H $\rightarrow$ H}  &
\textbf{L $\rightarrow$ H}  &
\textbf{H $\rightarrow$ L}
\\ \midrule
Remove & 3.12 & 3.23 & 1.78 & 4.03 \\
\bottomrule
\end{tabular}%
}
\caption{Average number of collisions in ground truth videos (i.e., vanilla) when different types of objects participate in first collision. ``$x \rightarrow y$'', where $x, y \in \{Light (L), Heavy (H)\}$, means that $x$ mass object collides with $y$ mass object.
}
\label{tab:first_col}
\end{table}

\section{Conclusion}
In this work, we present a new video question answering benchmark: CRIPP-VQA, for reasoning about the implicit physical properties of objects.
It contains novel tasks that require counterfactual reasoning and planning, over three hypothetical actions (i.e., remove, replace, and add). 
We evaluate state-of-the-art models on this benchmark and observe a significant performance gap between descriptive questions about visible properties and counterfactual and planning questions about implicit properties. 
We also show that models can learn the initial dynamics of object trajectories but they fail to detect subsequent collisions, which requires an understanding of relative mass. 
This result is positioned as a challenge for the V\&L community for building robust video understanding systems that can interact with language.

\section{Limitations}
While CRIPP proposes the implicit reasoning about intrinsic physical properties, it is limited to two physical properties (mass and friction).
However, even these fundamental properties are a big challenge for existing systems.
While other properties and complex dynamics can be considered, that is beyond the scope of this work.
Our benchmark is limited to a synthetic environment in blockworld, and we believe that future work should extend our work with real-world objects and backgrounds.

\section*{Acknowledgements}

This work was supported by NSF RI grants  \#1750082, \#1816039 and \#2132724, and the  DARPA GAILA ADAM project. 
The views and opinions of the authors expressed herein do not necessarily state or reflect those of the funding agencies and employers.
\bibliography{anthology,output,tgokhale}
\bibliographystyle{acl_natbib}

\appendix

\section*{Appendix}
\label{sec:appendix}

\section{Training details}
\label{sec:training}
We follow the standard training guidelines provided by the authors of each baseline papers. 
We train all systems on Quadro RTX 8000 GPUs. We train each model with a maximum of 200 epochs. 
And select the best model based on an average performance accuracy. 
We follow the below instructions to support each model which are MAC, HCRN, Aloe, and Aloe+BERT. 
For planning based task, we add four extra classifier heads on top of all models which predicts: 1) the type of the action, 2) an object on which action needs to be performed, 3) an object which needs to be added through replace or add action, and 4) relative direction of the object if we are adding a new object. 

\textbf{MAC:} We modify the public implementation of MAC from \url{https://github.com/rosinality/mac-network-pytorch} to adapt the video frames as input. 
We first resize the each 125 frames leading $(125, 3, 224, 224)$ video dimension. 
Later, we use ResNet101 to extract the features $(125, 512, 14, 14)$. 
After taking the channel-wise mean of features, we get the final video re-presentation of $(125, 14, 14)$ dimension matrix supportable for the rest of the pipeline. 
We also perform the necessary changes described for the planning task as well.

\textbf{HCRN:} As HCRN is the VideoQA model and official implementation is available at: \url{https://github.com/thaolmk54/hcrn-videoqa}, we use the source code as it is. Except we do important changes to do planning tasks.

\textbf{Aloe\textsuperscript{*}/Aloe\textsuperscript{*}+BERT:} We first reproduce the Aloe on PyTorch based on the architecture details from the research paper by \citep{ding2021attention} and their publicly available demo at \url{https://github.com/deepmind/deepmind-research/tree/master/object_attention_for_reasoning}. 
However, we use the code base from transformers\footnote{\url{https://github.com/huggingface/transformers}} library (as it is well tested and used across the industry and academia) and modify it to support the VideoQA in the same way as Aloe does. 
Our initial experiments on CLEVRER showed that Aloe cannot reproduce the results on CLEVRER with the specified set of architecture details and hyper-parameters from the original paper. 
Therefore, we do extensive experiments on Aloe architecture and hyper-parameter to reproduce similar results. 
After achieving a similar performance from the paper, we use this new reproducible Aloe architecture in our experiments. 
Table (\ref{tab:aloe_details}) shows the hyper-parameter details to reproduce the results. 
The Aloe\textsuperscript{*} source code from our experiments is available at \url{https://github.com/Maitreyapatel/CRIPP-VQA/}

 \begin{table}[!t]
\centering
\small
\begin{tabular}{@{}lr@{}}
\toprule
\textbf{Hyper-parameter} &
\textbf{Value} 
\\ \midrule
\# of layers & 28\\
\# of attention heads & 128\\
embedding size & 768\\
visual feature size & 512\\
text embedding size & 768\\
Batch Size for descriptive & 96\\
Batch Size for Counterfactual & 32\\
Batch Size for Planning & 16\\
Learning rate & 0.00005\\
Optimizer & RAdam\\ \bottomrule

\end{tabular}%
\caption{Aloe\textsuperscript{*}+BERT architecture and hyper-parameter details.}
\label{tab:aloe_details}
\end{table}
\begin{figure*}[t]
    \centering
    \includegraphics[width=\textwidth]{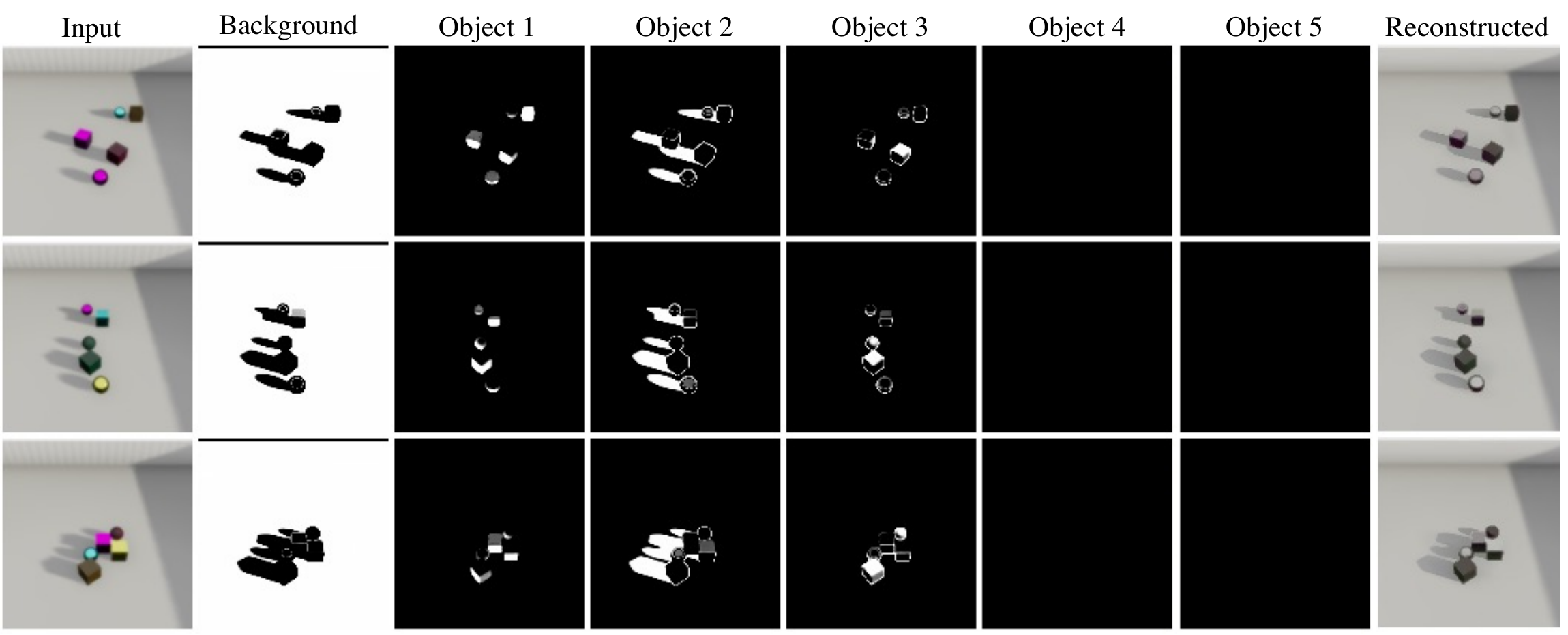}
    \caption{Example outputs of MONet-based scene decomposition failure cases. Left most images represents the input image. The Middle six images represent the predicted masks. And right most images represent the reconstructed input image by MONet.}
    \label{fig:monet_failure}
\end{figure*}

\section{Dataset Examples}
\begin{table*}[t]
\centering
\small
\begin{tabular}{@{}l|l@{}}
\toprule

\textbf{Question Type} & \textbf{Examples} \\
\midrule
\multirow{2}{*}{\textbf{Descriptive - Type 1}} & How many teal cardboard cube objects are there ?\\
& How many  cardboard sphere objects are static when video ends ? \\
\midrule
\multirow{2}{*}{\textbf{Descriptive - Type 2}} & Do teal cardboard cube objects exist in the video ?\\
& Do purple aluminium cube objects exist in the video ? \\
\midrule
\multirow{2}{*}{\textbf{Descriptive - Type 3}} & What is the color of the collidee of purple aluminium cube in collision number 1?\\
& What is the material of the collider of purple cardboard cube in collision number 2? \\
\midrule
\multirow{2}{*}{\textbf{Descriptive - Type 4}} & How many collisions are there between teal  sphere objects and teal aluminium  objects ?\\
& How many collisions are there between purple cardboard cube objects and teal objects ? \\
\midrule
\multirow{2}{*}{\textbf{Descriptive - Type 5}} & What is the maximum occurring shape of objects in the video ?\\
& What is the minimum occurring material of objects in the video ? \\
\midrule
\multirow{3}{*}{\textbf{Counterfactual - Remove}} & What will happen, if the  teal cardboard sphere is removed ? \\
& Choice: purple cardboard sphere would collide with purple cardboard cube \\
& Choice: teal cardboard cube would collide with purple cardboard cube \\
\midrule
\multirow{3}{*}{\textbf{Counterfactual - Replace}} & What will happen, if the  purple cardboard sphere is replaced by the purple aluminium sphere? \\
& Choice: purple aluminium sphere would collide with olive aluminium sphere \\
& Choice: teal cardboard sphere would collide with purple aluminium sphere \\
\midrule
\multirow{3}{*}{\textbf{Counterfactual - Add}} & What will happen, if the  purple cardboard sphere is added to the right of teal aluminium sphere? \\
& Choice: teal aluminium sphere would collide with purple cardboard cube \\
& Choice: olive aluminium cube would collide with teal aluminium sphere \\
\midrule
\multirow{2}{*}{\textbf{Planning}} & Make the collision between olive cardboard cube and olive aluminium sphere.\\
& Make the collision between teal cardboard sphere and olive cardboard sphere . \\


\bottomrule 
\end{tabular}%
\caption{
Examples of the CRIPP-VQA questions asked from different types of question categories as shown in Figure \ref{fig:cripp_data_stat}.
}
\label{tab:question_examples}
\end{table*}
The demo page contains several examples of the CRIPP-VQA dataset. 
Apart from that, Table \ref{tab:question_examples} shows the types of questions asked in different sub-categories of the QAs.

\section{MONet failure cases}
\label{sec:monet-appendix}
We discover that MONet-based unsupervised object decomposition does not function on complicated realistic visuals and that it is difficult to ensure that each object is decomposed on independent images/features. 
Here, we show three failure cases from the CRIPP-VQA. 
From the Figure \ref{fig:monet_failure}, we can observe that MONet is neither able to decompose the objects nor able to learn the color of the objects.
While MONet can learn the texture (i.e., metal or cardboard). 
As a result, we can see that the re-generated images lack greatly in terms the important features. 
Hence, we drop the MONet from the pipeline and adapt Mask R-CNN to work on our CRIPP dataset.

\section{Physical out-of-distribution results}
In this section, we provide accuracy tables for OOD evaluations. 
First, Table (\ref{tab:ood_mass8}) shows the performance of all models when the mass of few objects are changed (either increased or decreased to 8 from 2 or 14). 
Second, Table (\ref{tab:ood_nofric}) shows the results when the surface friction is removed. 
Third, Table (\ref{tab:ood_multiple}) shows the results where we have two objects initialized with fixed velocity creating more collisions. 
At last, Table (\ref{tab:ood_diff_highvel}) contains the results when we slightly increase the initial velocity of the object. 
Overall, we observe that for both counterfactual and planning tasks all model performs poorly.

\section{Neuro-symbolic methods} 
Recently, a lot of neuro-symbolic approaches are proposed for CLEVRER-like settings. 
For example, IEP \cite{johnson2017inferring}, NS-DR+ \citep{mao2019neuro}, are CPL \citep{chen2022comphy} proposed for physical reasoning. 
The goal of our study is to evaluate whether systems can learn the implicit relationship from counterfactual tasks. 
Symbolic approaches either require providing this implicit information or learning through a physics engine, which is not feasible for real-life situations. 
Therefore, in this study, we only consider neural models to evaluate their performance where learning implicit information is necessary.

\clearpage
 \begin{table*}[!t]
\centering\small
\begin{tabular}{@{}lccccccc@{}}
\toprule
\multicolumn{1}{l}{\multirow{2}{*}{\textbf{Model}}} &
\multicolumn{2}{c}{\multirow{1}{*}{\textbf{Remove}}}  &
\multicolumn{2}{c}{\multirow{1}{*}{\textbf{Replace}}}  &
\multicolumn{2}{c}{\multirow{1}{*}{\textbf{Add}}}  &
\multicolumn{1}{c}{\multirow{2}{*}{\textbf{Planning QA}}}
\\
\multicolumn{1}{c}{}&
\multicolumn{1}{c}{\textbf{PQ}}&
\multicolumn{1}{c}{\textbf{PO}}&
\multicolumn{1}{c}{\textbf{PQ}}&
\multicolumn{1}{c}{\textbf{PO}}&
\multicolumn{1}{c}{\textbf{PQ}}&
\multicolumn{1}{c}{\textbf{PO}}&
\multicolumn{1}{c}{}
\\ \midrule

Frequency & 0.00 & 50.27 & 0.00 & 50.00 & 0.00 & 50.00 & 21.00 \\
Random & 9.61 & 49.95 & 10.57 & 49.71 & 10.29 & 49.85 & 21.16 \\
Blind-BERT & 13.52 & 49.67 & 13.72 & 48.71 & 9.44 & 50.80 & 16.43 \\ \midrule 
MAC & 12.99 & 48.36 & 18.89 & 53.25 & 12.21 & 50.00 & 17.34 \\
HCRN & 18.15 & 55.47 & 20.08 & 56.84 & 14.03 & 54.58 & 43.94 \\
\midrule
Aloe\textsuperscript{*} & 20.46 & 57.00 & 12.98 & 52.05 & 12.90 & 54.93 & 46.07 \\ 
Aloe\textsuperscript{*}+BERT & 26.16 & 60.67 & 22.42 & 56.21 & 20.34 & 62.62 & 48.71 \\ \bottomrule 

\end{tabular}%
\caption{Performance evaluations when mass dist. is different than the training.}
\label{tab:ood_mass8}
\end{table*}
 \begin{table*}[!t]
\centering\small
\begin{tabular}{@{}lccccccc@{}}
\toprule
\multicolumn{1}{l}{\multirow{2}{*}{\textbf{Model}}} &
\multicolumn{2}{c}{\multirow{1}{*}{\textbf{Remove}}}  &
\multicolumn{2}{c}{\multirow{1}{*}{\textbf{Replace}}}  &
\multicolumn{2}{c}{\multirow{1}{*}{\textbf{Add}}}  &
\multicolumn{1}{c}{\multirow{2}{*}{\textbf{Planning QA}}}
\\
\multicolumn{1}{c}{}&
\multicolumn{1}{c}{\textbf{PQ}}&
\multicolumn{1}{c}{\textbf{PO}}&
\multicolumn{1}{c}{\textbf{PQ}}&
\multicolumn{1}{c}{\textbf{PO}}&
\multicolumn{1}{c}{\textbf{PQ}}&
\multicolumn{1}{c}{\textbf{PO}}&
\multicolumn{1}{c}{}
\\ \midrule

Frequency & 0.00 & 50.16 & 0.00 & 50.00 & 0.00 & 50.00 & 20.46 \\
Random & 10.41 & 50.51 & 3.92 & 50.42 & 6.58 & 49.93 & 20.49 \\
Blind-BERT & 10.86 & 49.90 & 11.90 & 49.77 & 13.96 & 50.80 & 12.34 \\ \midrule 
MAC & 11.6 & 50.30 & 14.23 & 49.82 & 7.86 & 50.53 & 14.07 \\
HCRN & 11.74 & 49.20 & 13.74 & 51.54 & 13.27 & 61.32 & 36.73 \\
\midrule
Aloe\textsuperscript{*} & 11.46 & 48.11 & 19.58 & 57.91 & 16.83 & 63.67 & 38.27 \\
Aloe\textsuperscript{*}+BERT & 7.21 & 48.85 & 24.87 & 55.93 & 18.37 & 64.66 & 43.67 \\ \bottomrule 

\end{tabular}%
\caption{Performance evaluations with zero surface friction.}
\label{tab:ood_nofric}
\end{table*}
 \begin{table*}[!t]
\centering
\small
\begin{tabular}{@{}lccccccc@{}}
\toprule
\multicolumn{1}{l}{\multirow{2}{*}{\textbf{Model}}} &
\multicolumn{2}{c}{\multirow{1}{*}{\textbf{Remove}}}  &
\multicolumn{2}{c}{\multirow{1}{*}{\textbf{Replace}}}  &
\multicolumn{2}{c}{\multirow{1}{*}{\textbf{Add}}}  &
\multicolumn{1}{c}{\multirow{2}{*}{\textbf{Planning QA}}}
\\
\multicolumn{1}{c}{}&
\multicolumn{1}{c}{\textbf{PQ}}&
\multicolumn{1}{c}{\textbf{PO}}&
\multicolumn{1}{c}{\textbf{PQ}}&
\multicolumn{1}{c}{\textbf{PO}}&
\multicolumn{1}{c}{\textbf{PQ}}&
\multicolumn{1}{c}{\textbf{PO}}&
\multicolumn{1}{c}{}
\\ \midrule

Frequency & 0.00 & 50.57 & 0.00 & 50.00 & 0.00 & 50.00 & 5.04 \\
Random & 5.09 & 49.92 & 4.76 & 50.75 & 9.00 & 49.37 & 8.15 \\
Blind-BERT & 12.64 & 49.70 & 12.70 & 49.64 & 5.82 & 51.58 & 7.57 \\ \midrule 
MAC & 14.87 & 50.84 & 13.78 & 52.99 & 12.17 & 50.33 & 8.51 \\
HCRN & 19.75 & 55.80 & 13.78 & 52.10 & 13.10 & 55.03 & 18.92 \\ \midrule

Aloe\textsuperscript{*} & 17.39 & 55.73 & 12.50 & 50.82 & 10.61 & 52.55 & 27.76 \\
Aloe\textsuperscript{*}+BERT & 12.81 & 55.10 & 18.37 & 52.54 & 21.10 & 60.86 & 25.95 \\ \bottomrule 

\end{tabular}%
\caption{Performance evaluations with multiple objects moving.}
\label{tab:ood_multiple}
\end{table*}
 \begin{table*}[!t]
\centering\small
\begin{tabular}{@{}lccccccc@{}}
\toprule
\multicolumn{1}{l}{\multirow{2}{*}{\textbf{Model}}} &
\multicolumn{2}{c}{\multirow{1}{*}{\textbf{Remove}}}  &
\multicolumn{2}{c}{\multirow{1}{*}{\textbf{Replace}}}  &
\multicolumn{2}{c}{\multirow{1}{*}{\textbf{Add}}}  &
\multicolumn{1}{c}{\multirow{2}{*}{\textbf{Planning QA}}}
\\
\multicolumn{1}{c}{}&
\multicolumn{1}{c}{\textbf{PQ}}&
\multicolumn{1}{c}{\textbf{PO}}&
\multicolumn{1}{c}{\textbf{PQ}}&
\multicolumn{1}{c}{\textbf{PO}}&
\multicolumn{1}{c}{\textbf{PQ}}&
\multicolumn{1}{c}{\textbf{PO}}&
\multicolumn{1}{c}{}
\\ \midrule

Frequency & 0.00 & 50.20 & 0.00 & 50.00 & 0.00 & 50.00 & 19.20 \\
Random & 10.93 & 49.47 & 3.66 & 50.12 & 6.99 & 49.83 & 19.53 \\
Blind-BERT & 10.02 & 50.27 & 16.06 & 52.17 & 5.73 & 51.82 & 13.68 \\ \midrule 
MAC & 10.76 & 49.80 & 15.09 & 51.44 & 6.34 & 48.66 & 11.54 \\
HCRN & 10.89 & 50.52 & 17.62 & 52.89 & 12.67 & 60.93 & 33.48 \\ \midrule

Aloe\textsuperscript{*} & 11.63 & 48.79 & 14.51 & 54.15 & 15.36 & 62.82 & 39.30 \\
Aloe\textsuperscript{*}+BERT & 6.43 & 48.63 & 29.53 & 61.37 & 15.82 & 65.02 & 42.10 \\ \bottomrule 

\end{tabular}%
\caption{Performance evaluations with higher initial velocity.}
\label{tab:ood_diff_highvel}
\end{table*}

\end{document}